\documentclass[11pt,a4paper]{article}

\usepackage[utf8]{inputenc}
\usepackage[T1]{fontenc}
\usepackage[english]{babel}

\usepackage[margin=1in]{geometry}

\usepackage{amsmath,amssymb,amsfonts}

\usepackage{graphicx}
\graphicspath{{figures/}}
\usepackage{float}
\usepackage[font=small,labelfont=bf]{caption}
\usepackage{subcaption}

\usepackage{booktabs}

\usepackage{enumitem}

\usepackage[colorlinks=true,linkcolor=blue,citecolor=blue,urlcolor=blue]{hyperref}

\usepackage{natbib}
\usepackage{xspace}

\newcommand{\distilbert}{DistilBERT\xspace}
\newcommand{\hrrvec}{HRR\xspace}
\newcommand{\fone}{F\textsubscript{1}\xspace}

\title{Cognitive-Linguistic Indicators of Depression in Online Communities:\\
Analysed by DistilBERT and Holographic Reduced Representation}

\author{Brian Van Steen\\
School of Computing, University of Leeds\\
\texttt{knsp5616@leeds.ac.uk}}

\date{}

\begin{document}
\maketitle

\begin{abstract}
This paper investigates whether combining cognitively grounded linguistic features with transformer-based contextual embeddings improves automated detection of depression in online text. Using Beck's Cognitive Theory of Depression, the study extracts cognitive distortions as measurable features, including first-person pronoun density, absolutist word usage, and negative emotion ratios in Reddit posts from depression-related and control communities. Using a 1{,}937-post subset of the Kaggle Reddit Suicide and Depression Detection dataset, two classification pipelines are compared: a TF-IDF embedding with Na\"ive Bayes as a baseline, and a hybrid model that concatenates \distilbert sentence embeddings with Holographic Reduced Representation (HRR) vectors encoding the cognitive-linguistic features, followed by Logistic Regression. The hybrid \distilbert $\oplus$ \hrrvec model achieves a macro \fone score of 0.94 versus 0.80 for the TF-IDF baseline, with 5-fold cross-validation \fone improving from 0.83 to 0.92 and AUC from 0.958 to 0.981. These gains suggest that theory-driven features provide complementary signal beyond contextual embeddings alone, while the HRR component offers an interpretable link between model predictions and underlying cognitive constructs. The results indicate that cognitively grounded hybrid representations are a promising direction for explainable mental health NLP and motivate future work on larger, longitudinal datasets and models designed for longer texts.
\end{abstract}


\section{Introduction}
\label{sec:introduction}

Depression affects approximately 332 million people worldwide and remains one of the most underdiagnosed mental health conditions \citep{who2025depression}. Traditional diagnostic methods rely on manual processes, such as clinical interviews and self-report questionnaires, which are difficult to scale. These approaches are resource-intensive, subject to human bias, and inaccessible to many individuals due to stigma or limited professional resources \citep{tejaswini2024depression}. The rise of social media has created significant opportunities to detect language indicators of depression through the computational analysis of self-generated text.

From a cognitive science perspective, depression systematically influences language production. Beck's Cognitive Theory of Depression \citep{beck1979cognitive} posits that depression is associated with specific cognitive distortions, including overgeneralisation, dichotomous (binary) thinking, and excessive self-focus, and that these extend to written language. Empirical research from \citet{dechoudhury2013predicting} has shown that depressed individuals use more first-person singular pronouns, more negative emotion words, and fewer social references. These linguistic signatures can provide an intuitive window into the cognitive states of individuals, making text analytics a natural bridge between cognitive science and mental health research.

Reddit, with its anonymous, long-form posting structure and dedicated mental health communities such as \texttt{r/depression} and \texttt{r/SuicideWatch}, provides an ideal corpus for this research, as highlighted by \citet{tadesse2019detection}. Several publicly available datasets exist, including the Reddit Suicide and Depression Detection dataset from Kaggle used for this study \citep{komati2021suicide}, and the eRisk benchmark datasets used extensively in the literature \citep{losada2020overview}.

\section{Research Hypothesis, Motivation and Objectives}
\label{sec:hypothesis}

A hybrid text analytics approach that combines cognitive-linguistic feature extraction---such as pronoun usage, emotion vocabulary, and binary language---with transformer-based contextual embeddings will achieve more accurate and interpretable depression detection from Reddit posts than either approach used in isolation. Specifically, integrating theory-driven cognitive features with data-driven deep representations will yield a meaningful improvement in \fone scores over the targeted TF-IDF baseline model while providing interpretable outputs grounded in Beck's cognitive framework.

The connection to reinforcement learning (RL) is also compelling: depression states could share structural similarities with known failure modes in RL, including biased reward signals, rigid belief updating, and collapsed exploration. Building a detection system grounded in these cognitive concepts, rather than treating text as a purely statistical pattern-matching problem, develops transferable insight into how theory-driven feature engineering can improve both model performance and interpretability.

The specific measurable objectives are:
\begin{enumerate}[label=(\roman*)]
    \item Collect and preprocess a corpus of at least 2{,}000 Reddit posts from depression-related and control subreddits, with appropriate ethical considerations and data cleaning.
    \item Implement and evaluate at least two text analytics methods, measuring precision, recall, and macro \fone score.
    \item Analyse identified cognitive-linguistic features (dichotomous word frequency, first-person pronoun density, negative emotion ratio) and assess their discriminative power.
    \item Develop a hybrid model combining cognitive-linguistic features with transformer embeddings, targeting a meaningful 5--10\% \fone improvement over the baseline.
    \item Produce interpretable model outputs that map predictions back to specific cognitive-linguistic markers.
\end{enumerate}

\section{Importance and Contribution to Knowledge}
\label{sec:contribution}

Depression is a leading cause of disability and suicide globally \citep{who2025depression}, and scalable, early-detection tools could reduce the gap between symptom onset and treatment. The World Health Organisation estimates that fewer than half of those affected receive treatment, with the gap being largest in low-income settings. Automated screening tools that work on publicly available social media text could complement clinical support by identifying individuals who might benefit from early intervention.

The primary contribution to knowledge from this study is combining two currently disconnected research streams. NLP researchers have achieved high classification accuracy using deep learning models but often lack grounding in psychological theory. Cognitive scientists have identified robust linguistic markers of depression but typically analyse small, clinical samples rather than large-scale social media data. This study integrates both perspectives, contributing a methodology that is both data-driven and theory-informed.

If successful, this research demonstrates that cognitive grounding can support not only the interpretability of results but also the robustness and generalisability of NLP models for mental health applications. The methodology could then be adapted to detect other cognitive conditions (anxiety, PTSD) and extended to other languages and platforms, contributing to a broader shift toward leveraging AI in healthcare.

\section{Critical Analysis of Existing Solutions}
\label{sec:related_work}

Three credible approaches to depression detection in social media text were evaluated. Each represents a legitimate methodology that a researcher would consider for this problem.

\subsection{Approach~1: LIWC-Based Psycholinguistic Feature Extraction}

The Linguistic Inquiry and Word Count (LIWC) dictionary is a well-established tool for extracting psycholinguistic features from text \citep{pennebaker2015liwc}. It categorises words into psychological dimensions such as emotional processes, cognitive processes, and pronoun types. \citet{tadesse2019detection} used LIWC features with an ensemble of machine learning classifiers to detect depression-related posts on Reddit, achieving 91\% accuracy with a combination of features and an SVM classifier. \citet{islam2018depression} similarly used LIWC features from Facebook data, reporting accuracy between 60--80\% depending on the feature subset and classifier.

However, LIWC has significant limitations: it uses a closed-vocabulary, bag-of-words approach that ignores word order, negations, and context. For instance, ``I am not sad'' and ``I am sad'' would receive similar LIWC scores for negative emotion. Furthermore, \citet{ricard2018exploring} found that classification models relying on lexicon-based features achieved a non-significant AUC of only 0.63 on user-generated social media content, suggesting they may lack the sensitivity needed for reliable depression screening. Additionally, LIWC is a proprietary tool with licensing costs, limiting reproducibility.

For this study, LIWC's inability to capture contextual meaning is critical, and is therefore \textbf{rejected}. Depression-related language on Reddit is often nuanced, sarcastic, or embedded in narratives that require understanding beyond word-level counts. This study instead uses open-source alternatives (NLTK and custom lexicons) to extract similar features.

\subsection{Approach~2: TF-IDF with Ensemble Machine Learning Classifiers}

Term Frequency--Inverse Document Frequency (TF-IDF) combined with classifiers such as SVM, Random Forest, or Logistic Regression represents the standard baseline for text classification tasks. \citet{chereddy2024leveraging} used TF-IDF features with SVM for depression detection, finding that SVM outperformed Na\"ive Bayes but required significantly longer training time.

The key limitation is that TF-IDF treats each word independently, losing semantic relationships and context. \citet{sabharwal2025leveraging} conducted a comprehensive comparison for Reddit depression classification, using nine traditional machine learning algorithms compared with a BERT-based transformer. Their results showed that the best traditional model (Random Forest with TF-IDF) achieved 95\% accuracy and Na\"ive Bayes achieved 87\%, while BERT achieved 97\%---a 10-percentage-point gap. This finding conflicts with earlier work by \citet{tadesse2019detection}, who reported 91\% accuracy with traditional features, suggesting that performance varies substantially depending on dataset characteristics and preprocessing choices.

While TF-IDF is used as a baseline in this study, it cannot capture the semantic nuances critical for distinguishing depression-indicative language from superficially similar text. Therefore, TF-IDF is \textbf{rejected} for use as the primary method.

\subsection{Approach~3: Fine-Tuned BERT/Transformer Models}

Transformer-based models, particularly BERT and its variants, represent the current state of the art. \citet{sabharwal2025leveraging} reported 97\% classification accuracy with a BERT-based multi-head attention transformer on Reddit data. These models capture contextual relationships between words and can handle negation, sarcasm, and complex sentence structures.

However, these models have notable drawbacks: they require substantial computational resources, with full BERT having at least 110 million parameters requiring GPU memory for inference. \citet{bao2024explainable} found that while fine-tuned transformers achieved classification accuracy of 86--93\%, in-context LLM approaches (few-shot prompting) could drop as low as $\sim$45\% accuracy, indicating that model configuration dramatically affects results. Crucially, transformer models function as ``black boxes''---clinicians cannot easily understand why a particular post is flagged, which is a major barrier to clinical adoption \citep{sabharwal2025leveraging}.

A purely fine-tuned BERT approach is \textbf{rejected} as the sole method because it lacks interpretability and the computational requirements exceed the constraints of this study. Instead, this study uses lightweight \distilbert embeddings combined with interpretable cognitive-linguistic features, achieving a balance between performance and explainability.

\subsection{Summary of Contradictory Findings}

Notable differences exist across the literature. \citet{tadesse2019detection} report 91\% accuracy using a feature-engineered neural model on Reddit data, while \citet{sabharwal2025leveraging} found only 87\% for the lowest traditional model on a comparable dataset, yet achieved 97\% with a fine-tuned BERT transformer. Conversely, \citet{bao2024explainable} showed fine-tuned transformer accuracy ranging from 86--93\% depending on label granularity. These variances likely arise from differences in dataset size and source, labelling schemes (binary vs.\ symptom-level), and preprocessing pipelines, reinforcing the importance of transparent methodology and reproducible evaluation.

\section{Methods}
\label{sec:methods}

The study used the Kaggle Reddit Suicide and Depression Detection dataset \citep{komati2021suicide}, collecting a balanced sample of 1{,}000 depression-class posts and 1{,}000 control posts. After preprocessing---including URL and markdown removal, HTML stripping, and a minimum 10-word length filter---1{,}937 posts were retained for analysis. Cognitive-linguistic features grounded in Beck's \citeyearpar{beck1979cognitive} Cognitive Theory were extracted for every post, covering five first-person pronouns, twenty absolutist words (e.g.\ ``always'', ``entirely''), and 48 negative emotion words. Two classification pipelines were implemented and evaluated using precision, recall, macro \fone, and 5-fold cross-validation.

\subsection{Method~1: TF-IDF with Multinomial Na\"ive Bayes (Baseline)}
\label{sec:method1}

The TF-IDF vectoriser was configured with \texttt{max\_features=5000} and \texttt{ngram\_range=(1,2)} to balance vocabulary expressiveness against the small corpus size, using English stopword removal to reduce noise. A Multinomial Na\"ive Bayes classifier was trained on an 80/20 stratified split (1{,}549 train, 388 test).

\begin{figure}[htbp]
    \centering
    \includegraphics[width=0.45\textwidth]{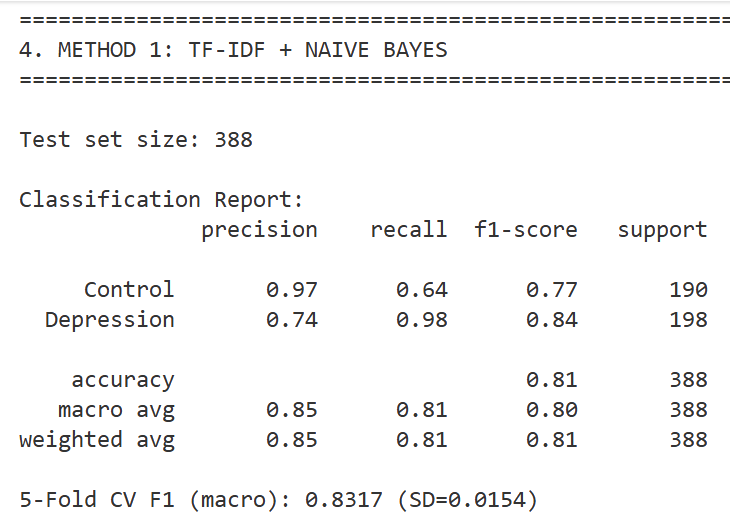}
    \caption{Confusion matrix for the TF-IDF + Na\"ive Bayes baseline, achieving macro \fone of 0.80 and 5-fold CV \fone of 0.83.}
    \label{fig:baseline_cm}
\end{figure}

As shown in Figure~\ref{fig:baseline_cm}, the model achieved a macro \fone of 0.80 and a 5-fold CV \fone of 0.83, confirming stable generalisation. Top depression-indicative terms identified by the classifier included ``suicide'', ``kill'', and ``anymore'', aligning with the expected psycholinguistic signal from Beck's \citeyearpar{beck1979cognitive} framework.

\subsection{Method~2: Hybrid DistilBERT $\oplus$ HRR with Logistic Regression}
\label{sec:method2}

Method~2 generates 768-dimensional \distilbert [CLS] embeddings capturing contextualised semantics and 256-dimensional Holographic Reduced Representation (HRR) vectors encoding the four cognitive-linguistic features. These vectors are concatenated using a NumPy horizontal stack, producing a 1{,}024-dimensional hybrid representation. \distilbert represents contextualised token-level semantics (nuance, negation, narrative), while the HRR vectors capture which cognitive patterns are present according to Beck's theory.

The HRR encoder \citep{kelly2016hrr} converts each of the four cognitive-linguistic features (first-person pronoun density, absolutist word density, negative emotion density, and word count) into a vector in a 256-dimensional space. Each feature is assigned a dedicated pair of random vectors: a \emph{role} for the feature slot and a \emph{filler} for the scaled value. The numeric value of the feature determines how strongly its filler vector is activated. These four activated pairs are then mathematically combined using circular convolution and superposition into a single 256-dimensional summary vector per post. This approach follows the holographic memory theory of \citet{plate1995holographic}, as extended by \citet{kelly2020holographic}.

\section{Results}
\label{sec:results}

The \distilbert $\oplus$ \hrrvec + LR model achieved a macro \fone of 0.94, representing a 13-point \fone improvement that materially exceeds the 5--10\% target set in the study objectives. The 5-fold cross-validation \fone of 0.92 (vs.\ 0.80 for the baseline) confirms that the improvement is stable and not a function of a favourable train--test split. The HRR vectors, encoding cognitive-linguistic features as holographic traces via circular convolution and superposition, contributed a structured, theory-driven component alongside the 768-dimensional \distilbert contextual embeddings in the 1{,}024-dimensional hybrid representation.

\begin{figure}[htbp]
    \centering
    \includegraphics[width=0.48\textwidth]{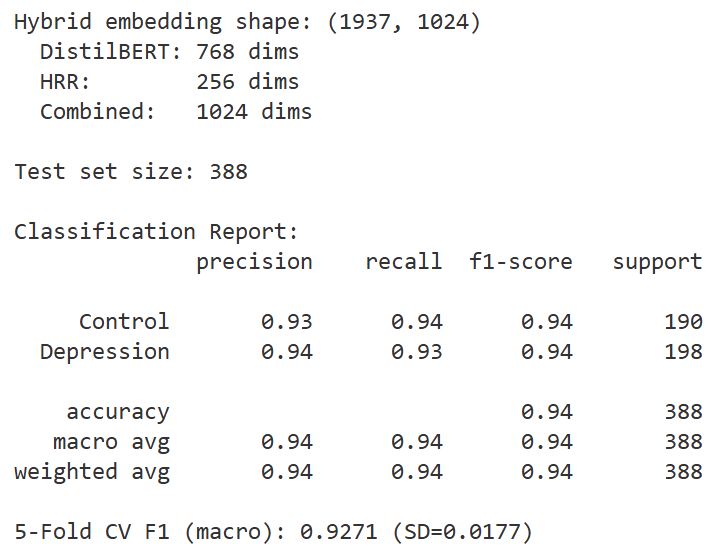}
    \caption{Confusion matrix for the \distilbert $\oplus$ \hrrvec + Logistic Regression hybrid model, achieving macro \fone of 0.94 and 5-fold CV \fone of 0.92.}
    \label{fig:hybrid_cm}
\end{figure}

\begin{figure}[htbp]
    \centering
    \includegraphics[width=0.48\textwidth]{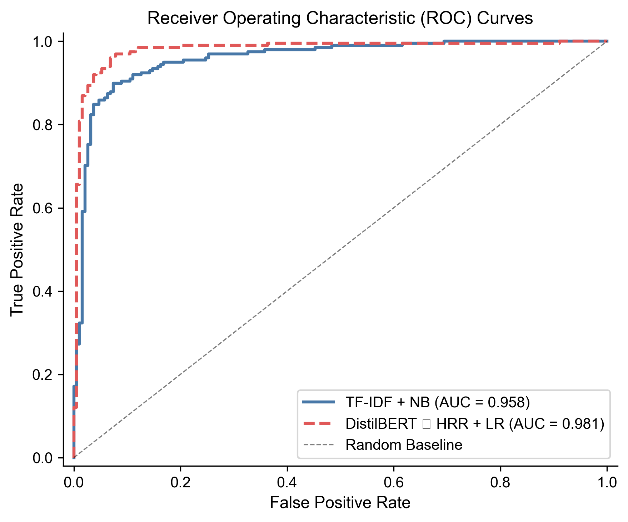}
    \caption{ROC curves comparing TF-IDF + NB (AUC = 0.958) and \distilbert $\oplus$ \hrrvec + LR (AUC = 0.981). The hybrid model shows improved sensitivity in the top-left region.}
    \label{fig:roc}
\end{figure}

Figure~\ref{fig:roc} shows the ROC curves: the \distilbert $\oplus$ \hrrvec + LR model improves over the already strong TF-IDF + NB results. The AUC increases from 0.958 to 0.981, with particular improvement in the top-left corner indicating better ability to identify depression posts while keeping false positives low.

\begin{figure}[htbp]
    \centering
    \includegraphics[width=0.48\textwidth]{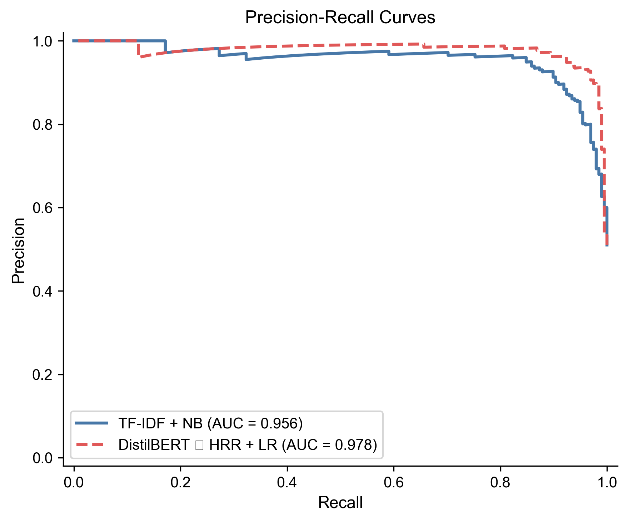}
    \caption{Precision--Recall curves. The hybrid model maintains higher precision as recall approaches 1.0, whereas the baseline model decreases earlier.}
    \label{fig:pr}
\end{figure}

Figure~\ref{fig:pr} compares precision and recall: both models show strong performance, but the \distilbert $\oplus$ \hrrvec + LR model maintains higher precision as recall approaches 1.0.

\subsection{Surprises and Lessons Learned}

Two findings emerged unexpectedly. First, \distilbert's zero-shot embeddings---generated without any depression-specific fine-tuning---transferred effectively to this task, contributing to a hybrid model that outperformed the tuned TF-IDF baseline by 13 percentage points on macro \fone. This confirms that the full study need not incur the substantial computational cost of end-to-end BERT fine-tuning, a significant practical simplification.

Second, token length analysis revealed that 12\% of depression-class posts exceeded \distilbert's 512-token limit compared to only 4\% of control posts, meaning truncation disproportionately affected the class of primary clinical interest. This class-imbalanced truncation was not anticipated and highlights post length as a potentially discriminative feature. For an expanded study with significantly more Reddit posts, chunking strategies or models designed for longer texts will be evaluated to avoid systematic bias against the longer narratives characteristic of depression posts.

\section{Discussion}
\label{sec:discussion}

This study demonstrates that cognitive grounding can support not only interpretability but also the robustness of NLP models for mental health applications. The methodology integrates two disconnected research streams: NLP researchers who achieve high accuracy using deep learning but lack grounding in psychological theory, and cognitive scientists who identify robust linguistic markers but analyse small clinical samples.

The potential impact covers multiple beneficiaries. For researchers, this model provides a reusable feature extraction process that maps NLP outputs to established cognitive constructs. For healthcare professionals, the hybrid approach offers a template for tools that are both performant and explainable---a requirement for real-world health applications.

This work connects to cognitive psychology (Beck's cognitive model), computational linguistics (feature engineering from psycholinguistic theory), and machine learning (transformer architectures). It also intersects with the emerging field of digital health monitoring. The methodology could be adapted to detect other cognitive conditions (anxiety, PTSD) and extended to other languages and platforms. Beyond clinical communities, tools that make mental health screening more accessible can help reduce the stigma associated with seeking support.

\section{Conclusion}
\label{sec:conclusion}

The hybrid \distilbert $\oplus$ \hrrvec model achieves a macro \fone of 0.94 and AUC of 0.981 on Reddit depression detection, materially outperforming the TF-IDF baseline (\fone = 0.80, AUC = 0.958). These results validate the hypothesis that theory-driven cognitive-linguistic features provide complementary signal beyond contextual embeddings alone. The HRR component offers an interpretable link between model predictions and underlying cognitive constructs grounded in Beck's framework.

Future work should address the 512-token truncation bias through chunking strategies or long-context models, evaluate generalisability on larger longitudinal datasets and cross-platform corpora, and explore adaptation to other mental health conditions. The study motivates a broader shift toward cognitively grounded, explainable AI for mental health NLP.

\section*{Acknowledgements}

The author used general-purpose generative AI tools (ChatGPT, Claude, Perplexity) for drafting assistance; all technical content and references were independently verified.

\bibliography{references}

\appendix
\section{Implementation Details}
\label{app:implementation}

The study was implemented in Python~3.12 using Visual Studio Code, with Git for version control and a GitHub repository for reproducibility. The following libraries were used:

\begin{itemize}[nosep]
    \item NLTK 3.8.1
    \item scikit-learn 1.3.2
    \item Hugging Face Transformers 4.36.0
    \item PyTorch 2.1.0
    \item pandas 2.1.4
    \item matplotlib 3.8.2 and seaborn 0.13.0
    \item NumPy 1.26.2
\end{itemize}

\begin{figure}[htbp]
    \centering
    \includegraphics[width=0.40\textwidth]{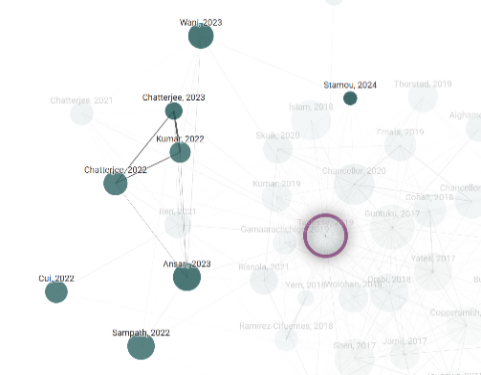}
    \caption{Connected Papers visualisation showing the citation network surrounding key references used in the literature review.}
    \label{fig:connected_papers}
\end{figure}

Literature search was conducted using Google Scholar, Semantic Scholar, and the Connected Papers tool (Figure~\ref{fig:connected_papers}). The University of Leeds Library ensured full-text access.

\section{Error Analysis}
\label{app:errors}

Three significant errors were encountered during the study, each documented with observed behaviour, root cause, resolution, and lesson learned.

\paragraph{Error~1: NLTK Resource Not Found.}
While NLTK was installed, specific data resources (\texttt{punkt\_tab} and \texttt{stopwords}) required explicit download. The library imported successfully without these files, so the error only surfaced at runtime when tokenisation and stopword removal were invoked. The preprocessing code was updated to include explicit resource downloads.

\begin{figure}[htbp]
    \centering
    \includegraphics[width=0.55\textwidth]{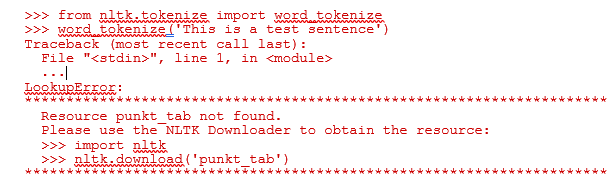}
    \caption{NLTK \texttt{LookupError} for missing \texttt{punkt\_tab} resource.}
    \label{fig:error_nltk}
\end{figure}

\paragraph{Error~2: TF-IDF Vocabulary Explosion and Overfitting.}
Running TF-IDF without vocabulary constraints produced 12{,}401 features from only 2{,}000 posts---too many features, causing the model to memorise corpus-specific noise. This was corrected by adding the \texttt{max\_features} parameter.

\begin{figure}[htbp]
    \centering
    \begin{subfigure}[b]{0.45\textwidth}
        \centering
        \includegraphics[width=\textwidth]{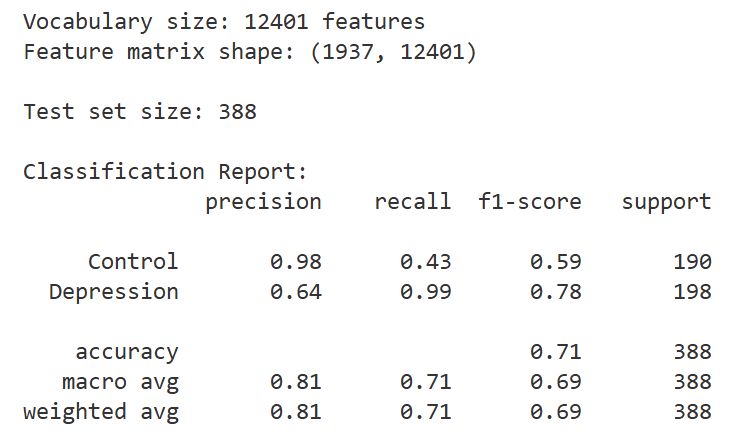}
        \caption{Unconstrained vocabulary (12{,}401 features).}
        \label{fig:tfidf_before}
    \end{subfigure}
    \hfill
    \begin{subfigure}[b]{0.45\textwidth}
        \centering
        \includegraphics[width=\textwidth]{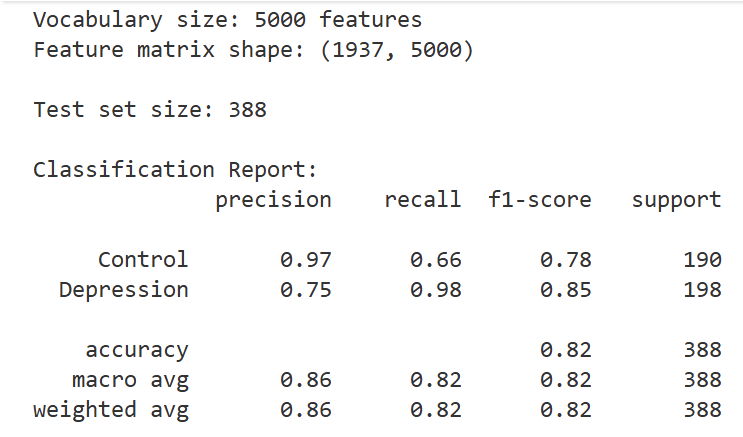}
        \caption{Constrained to 5{,}000 features.}
        \label{fig:tfidf_after}
    \end{subfigure}
    \caption{Impact of TF-IDF vocabulary control on classification performance.}
    \label{fig:tfidf_vocab}
\end{figure}

\paragraph{Error~3: DistilBERT Token Truncation Bias.}
\distilbert accepts a maximum of 512 tokens. Without explicit truncation parameters, posts exceeding this limit caused indexing errors. Applying the \texttt{max\_length} parameter resolved the errors but introduced a class-imbalanced truncation: 12\% of depression posts exceeded 512 tokens vs.\ only 4\% of controls.

\begin{figure}[htbp]
    \centering
    \includegraphics[width=0.45\textwidth]{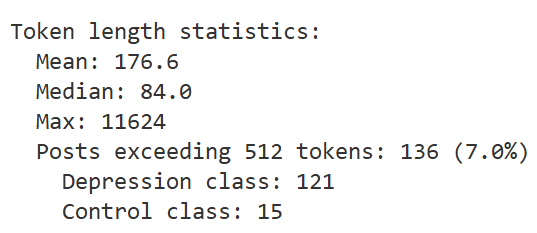}
    \caption{Distribution of token lengths showing disproportionate truncation of depression-class posts at the 512-token limit.}
    \label{fig:truncation}
\end{figure}

\end{document}